%% file: main.tex
\documentclass[10pt,twocolumn,letterpaper]{article}

\usepackage{cvpr}              
\input{preamble}
\usepackage{algorithm}
\usepackage{algorithmic}
\usepackage{pgfplots}
\pgfplotsset{compat=1.18}
\usepackage{subcaption}
\usepackage{amsmath}
\definecolor{cvprblue}{rgb}{0.21,0.49,0.74}
\usepackage[pagebackref,breaklinks,colorlinks,allcolors=cvprblue]{hyperref}

\def\paperID{EarthVision-35} 
\def\confName{EarthVision}
\def\confYear{2026}

\title{CraterBench-R: Instance-Level Crater Retrieval for Planetary Scale}

\author{Jichao Fang\thanks{Corresponding author.}\\
Northern Illinois University\\
{\tt\small jfang2@niu.edu}
\and
Lei Zhang\\
Northern Illinois University\\
{\tt\small zhanglei@niu.edu}
\and
Michael Phillips\\
University of Arizona\\
{\tt\small phillipsm@arizona.edu}
\and
Wei Luo\\
Northern Illinois University\\
{\tt\small wluo@niu.edu}
}

\begin{document}
\maketitle

\input{sec/0_abstract}

\input{sec/1_intro}
\input{sec/1b_related}
\input{sec/2_methods}

\input{sec/2b_method}
\input{sec/3_experiments}
{
    \small
    \bibliographystyle{ieeenat_fullname}
    \bibliography{main}
}

\input{sec/X_suppl}

\end{document}

%% file: preamble.tex
\DeclareMathOperator*{\argmax}{arg\,max}


%% file: sec/0_abstract.tex
\begin{abstract}
Impact craters are a cornerstone of planetary surface analysis. However, while most deep learning pipelines treat craters solely as a detection problem, critical scientific workflows—such as catalog deduplication, cross-observation matching, and morphological analog discovery—are inherently retrieval tasks. To address this, we formulate crater analysis as an instance-level image retrieval problem and introduce CraterBench-R, a curated benchmark featuring ~25k crater identities with multi-scale gallery views and manually verified queries spanning diverse scales and contexts. Our baseline evaluations across various architectures reveal that self-supervised Vision Transformers (ViTs), particularly those with in-domain pretraining, dominate the task, outperforming generic models with significantly more parameters. Furthermore, we demonstrate that retaining multiple ViT patch tokens for late-interaction matching dramatically improves accuracy over standard single-vector pooling. However, storing all tokens per image is operationally inefficient at a planetary scale. To close this efficiency gap, we propose instance-token aggregation, a scalable, training-free method that selects $K$ seed tokens, assigns the remaining tokens to these seeds via cosine similarity, and aggregates each cluster into a single representative token. This approach yields substantial gains: at $K=16$, aggregation improves mAP by +17.9 points over raw token selection, and at $K=64$, it matches the accuracy of using all 196 tokens with significantly less storage. Finally, we demonstrate that a practical two-stage pipeline—single-vector shortlisting followed by instance-token reranking—recovers 89–94\% of the full late-interaction accuracy while searching only a small candidate set. The benchmark is publicly available at \url{https://hf.co/datasets/jfang/CraterBench-R}.
\end{abstract}

%% file: sec/1_intro.tex
\section{Introduction}
\label{sec:intro}
Impact craters are both a dominant geomorphic element and a key quantitative tool for planetary science \cite{melosh1989impact}. Crater size--frequency distributions underpin surface-age estimation \cite{crater1979standard, neukum2001cratering, fassett2016analysis}, while crater morphology (e.g., rim sharpness, terracing, ejecta texture, infill) encodes information about target properties \cite{pike1980control}, degradation history \cite{soderblom1970model, fassett2014crater}, and resurfacing processes \cite{strom1994global}. The scale of modern orbital imaging, however, has outpaced manual analysis: global mosaics contain millions of crater-like structures spanning orders of magnitude in diameter, illumination, and preservation state. Recent deep learning efforts have therefore focused primarily on \emph{crater detection}---predicting crater locations and sizes from images or digital elevation models (DEMs)---with substantial progress on the Moon and Mars \cite{silburt2019lunar,delatte2019segmentation,lee2019automated}. Detection is indispensable, but its outputs---locations and diameters---do not provide the visual representations required by mapping workflows that rely on \emph{association}: deduplicating overlapping detections across image footprints \cite{mu2023yolo,tewari2023craterreview}, linking views of the same physical crater across scale and context \cite{yu2014matching,yang2023cratermatching,cheng2013registration}, and grouping candidates by morphology or degradation state \cite{liu2024global,alidib2021automated}.

These operational needs motivate a complementary view: crater analysis as \emph{instance-level image retrieval}.
Given a query view of a crater, the system retrieves other images depicting the same physical instance (instance retrieval) or morphologically similar craters (analog search).
We focus on instance retrieval as the first benchmark task because it admits objective, unambiguous ground truth; analog retrieval, while scientifically valuable, requires subjective similarity judgments that complicate evaluation. 

This setting is nonetheless profoundly challenging in orbital imagery. Martian craters exhibit extreme visual complexity due to diverse degradation states (e.g., pristine vs. heavily eroded rims), infilling mechanisms (sand dunes, dust, lava), and radical illumination changes between orbital passes. This complexity creates severe structural and photometric variations, compounded by large scale shifts and scarce repeat observations in our dataset.

To enable a systematic study, we introduce \textbf{CraterBench-R}, a curated instance-retrieval benchmark on Mars CTX imagery with $\sim$25k crater identities and manually verified multi-view queries designed to stress scale and context variation. Through a systematic evaluation of state-of-the-art vision backbones on this benchmark, we identify a critical representation bottleneck in applying foundation models to planetary science retrieval. First, we find that collapsing Vision Transformer (ViT) patch tokens into a single global descriptor (e.g., via CLS or GeM pooling) heavily compresses spatial detail, resulting in a low accuracy ceiling. The standard computer vision reflex to this problem would be to apply supervised metric learning to learn a better global embedding. However, on CraterBench-R, standard supervised metric fine-tuning with three widely used losses consistently \emph{degrades} retrieval, including late-interaction accuracy, likely because the limited number of distinct views per crater (two per identity) provides insufficient positive diversity for effective representation learning. Consequently, we rely on the \emph{frozen, multi-token} patch representations. While retaining all $N{=}196$ tokens and matching via late interaction improves accuracy, dense-token late interaction is operationally infeasible as a first-stage search method at planetary scale due to its storage footprint and query cost.

To bridge the gap between single-vector efficiency and dense-token accuracy, we propose \emph{instance-token aggregation}, which is a scalable, entirely training-free pipeline that elegantly compresses dense ViT patches into $K \ll 196$ highly discriminative instance tokens. By prioritizing salient spatial seeds and aggregating surrounding tokens via nearest-neighbor residual assignment, our method preserves local crater morphology without the blurring effect of classical $K$-means centroids. Because it operates entirely on frozen features, it circumvents the pitfalls of fine-tuning on limited planetary views. The resulting instance tokens are matched via late interaction, achieving near-dense token accuracy at a fraction of the storage footprint and search time.

Beyond crater retrieval, this work addresses a general challenge in GeoAI: scaling instance-level retrieval over large embedding corpora produced by geo-foundation models. The methodology we develop---late-interaction matching, deterministic post-hoc token compression, and two-stage coarse-to-fine search---is domain-agnostic and directly applicable to Earth observation tasks such as change detection, scene deduplication, and geographic localization. We use Mars as a testbed not only because it isolates the retrieval challenge under extreme domain shift (free from confounders like seasonal variation, cloud cover, or label noise that complicate terrestrial benchmarks), but also because it serves as an excellent proving ground for the robustness of Earth Observation models when confronted with unfamiliar, visually complex geographies. The resulting pipeline generalizes to any setting where frozen ViT features must be matched efficiently at scale.

\paragraph{Contributions.}
We make four contributions:
\begin{itemize}
    \item \textbf{Task + benchmark (CraterBench-R).}
    We formulate crater analysis as \emph{instance-level image retrieval} and introduce {CraterBench-R},
    a curated Mars CTX benchmark with $\sim$25k crater identities, 50k gallery images (multi-scale context),
    and 5k manually verified multi-view queries, along with an evaluation protocol.

    \item \textbf{Baseline diagnosis for planetary retrieval.}
    Across 30 frozen backbones, we show that (i) single-vector pooling imposes a low accuracy ceiling
    and (ii) supervised metric-learning fine-tuning degrades retrieval in this regime,
    while token-level matching yields large gains.

    \item \textbf{Instance-Token Aggregation (training-free).}
    We propose a deterministic, post-hoc compression scheme that converts frozen ViT
    patch tokens into $K\!\ll\!196$ \emph{instance tokens} via seed selection and nearest-neighbor residual assignment,
    preserving local morphology for late-interaction matching without any learned parameters.

    \item \textbf{Planetary-scale retrieval pipeline.}
    We demonstrate a practical two-stage system---single-vector FAISS shortlisting followed by instance-token reranking---that recovers 89--94\% of exhaustive late-interaction accuracy at $S{=}100$ (and up to $\sim$96\% at $S{=}500$),
    with millisecond-scale per-query latency and robustness to compression.
\end{itemize}

%% file: sec/1b_related.tex
\section{Related Work}
\label{sec:related}

\paragraph{Crater analysis, catalogs, and similarity-based crater studies.}
Deep learning for crater analysis spans detection/segmentation, morphology
estimation, and catalog construction on orbital imagery and DEMs
\cite{delatte2019segmentation,giannakis2023crater,lee2019automated,
martinez2025crater,silburt2019lunar,zhao2024crater}. These pipelines
typically output geometric parameters (e.g., centers and diameters) that are
well suited for counting and mapping, but do not directly provide visual
representations for comparing craters across observations or contexts.
Closer to our goal, Ali-Dib et al.~\cite{alidib2021automated} learn crater
shape descriptors and demonstrate that similarity-based reasoning enables
morphological grouping and large-sample analysis. However, prior work does
not formulate crater analysis as \emph{instance-level image retrieval}---matching
 crater identities across views and scales, nor provide a benchmark
to evaluate retrieval systems under controlled protocols.

\paragraph{Instance retrieval and ViT representations under resource constraints.}
Instance retrieval traditionally relies on compact global descriptors obtained by
pooling CNN or ViT features---e.g., GeM~\cite{radenovic2017gem}, R-MAC~\cite{tolias2016rmac},
or learned aggregation such as NetVLAD~\cite{arandjelovic2016netvlad}.
Local feature pipelines (DELF~\cite{noh2017delf}, DELG~\cite{cao2020delg}) retain spatial
structure for verification and reranking, but require keypoint detection and matching.
Recent work has shown that self-supervised ViT features, particularly from the DINO family
\cite{caron2021dino,oquab2023dinov2,simeoni2025dinov3}, exhibit strong zero-shot retrieval behavior,
partly due to emergent patch-level correspondence that makes token-level matching attractive.
Exploiting this structure raises a representation dilemma: single-vector pooling discards the
spatial detail that makes ViT features powerful, while retaining all patch tokens is expensive for
large-scale search. Prior work on token efficiency in ViTs---including pruning~\cite{rao2021dynamicvit},
merging~\cite{bolya2023tome}, and grouping-based aggregation~\cite{xu2022gtp}---primarily targets
throughput within the transformer forward pass for recognition or dense prediction, rather than
retrieval descriptor quality on frozen features. We address the complementary retrieval-time problem:
compressing post-hoc frozen patch tokens into compact multi-vector descriptors suitable for efficient
late-interaction reranking.

\paragraph{Scalable multi-vector retrieval and late interaction.}
Late interaction, introduced by ColBERT~\cite{khattab2020colbert}, scores queries against candidates
via token-level similarity aggregation and can outperform single-vector matching when fine-grained
correspondence matters. Scaling late interaction has motivated retrieval system engineering that reduces
multi-vector storage and accelerates matching: ColBERTv2~\cite{santhanam2022colbertv2} couples residual
compression with denoised supervision to shrink the per-document footprint, while PLAID~\cite{santhanam2022plaid}
accelerates search via centroid-based candidate generation and pruning.
Large-scale retrieval further relies on approximate nearest-neighbor indexing and vector compression
(e.g., product quantization and billion-scale GPU search) \cite{jegou2011pq,johnson2019faiss}.
Our work brings these ideas to planetary imagery: we study late interaction over ViT patch tokens on a
new crater benchmark, and introduce a training-free instance-token aggregation scheme that enables a
two-stage coarse-to-fine pipeline---compact indexing followed by budgeted late-interaction reranking---at
catalog scale.

%% file: sec/2_methods.tex
\section{CraterBench-R: Dataset and Evaluation Protocol}
\label{sec:dataset}

\paragraph{Overview} CraterBench-R is a curated \emph{instance-level} crater retrieval benchmark built from Mars CTX mosaic 
and crater identities from the Robbins catalog \cite{robbins2012new}.
It contains 25,000 crater identities with a 50,000-image gallery (two canonical context  per crater: 2$\times$ and 3$\times$ diameter crops),
and a manually verified query set of 5,000 images (1,000 craters $\times$ 5 views) designed to stress \emph{scale, context, and photometric variation}
(Fig.~\ref{fig:dataset-view}).
Unlike detection datasets, CraterBench-R is organized by crater \emph{identity} and evaluated under a retrieval protocol with  relevance definitions.

\subsection{Task Definition}
Real-world planetary retrieval encompasses both analog search (discovering morphologically similar  craters) and instance retrieval (linking observations of the exact same physical crater). While analog search is critical for comparative geomorphology, analog similarity is inherently subjective and lacks unambiguous ground truth. Therefore, CraterBench-R rigorously formalizes the core visual challenge as strict instance-level identity matching. Operating on the premise that an algorithm must successfully re-identify the exact same physical crater under severe scale and context shifts before it can reliably cluster morphological analogs, we define our task as follows: Given a query image $q$ depicting a crater, the goal is to retrieve from a gallery $\mathcal{G}$ the images that depict the same physical  instance. Each image is labelled with a crater catalog identifier from the Robbins Mars crater database~\cite{robbins2012new} (385{,}049 craters, $D{\geq}1$\,km, with metadata including diameter, ellipticity, ejecta morphology, and rim/floor degradation state).

Because image footprints of small craters can overlap those of nearby neighbours, we adopt \emph{cluster-tolerant} relevance: each query $q$ carries a set of co-visible crater IDs $\mathcal{I}(q)$, and a gallery image $g$ is considered relevant if it shares at least one ID with $\mathcal{I}(q)$.
In practice, 94\% of query craters map to a single gallery identity ($|\mathcal{I}(q)|{=}1$); the remaining 6\% include 2--10 co-visible neighbours, so the protocol is effectively a general one-to-one matching task.

\subsection{Dataset Construction}
\label{sec:dataset_construction}

We construct a retrieval benchmark from 25{,}000 crater identities (6.5\% of the full catalog) distributed across 8 Mars Chart (MC) quadrangles in the equatorial belt (latitude $\pm 30^\circ$): MC-02 through MC-03, MC-06 through MC-07, MC-10 through MC-11, and MC-14 through MC-15. All the imagery is from fully controlled CTX mosaic~\cite{robbins2023ctxmosaic}.
Crater diameters range from 1.0 to 401\,km (median 1.5\,km); 69\% are smaller than 2\,km.
Of the 25{,}000 craters, 10.3\% have classified ejecta morphology and degradation states span all four levels approximately equally.

\paragraph{Gallery.}
For each crater ID we generate two canonical views at different spatial extents (``$2\times$'' and ``$3\times$'' crater diameter context crops), yielding 50{,}000 gallery images. We include two canonical gallery views (2$\times$, 3$\times$ diameter context) to explicitly evaluate robustness to context change while keeping the identity fixed. We apply robust filtering to remove missing tiles, corrupted imagery, and extreme low-quality samples.

\begin{figure}
    \centering
    \includegraphics[width=1\linewidth]{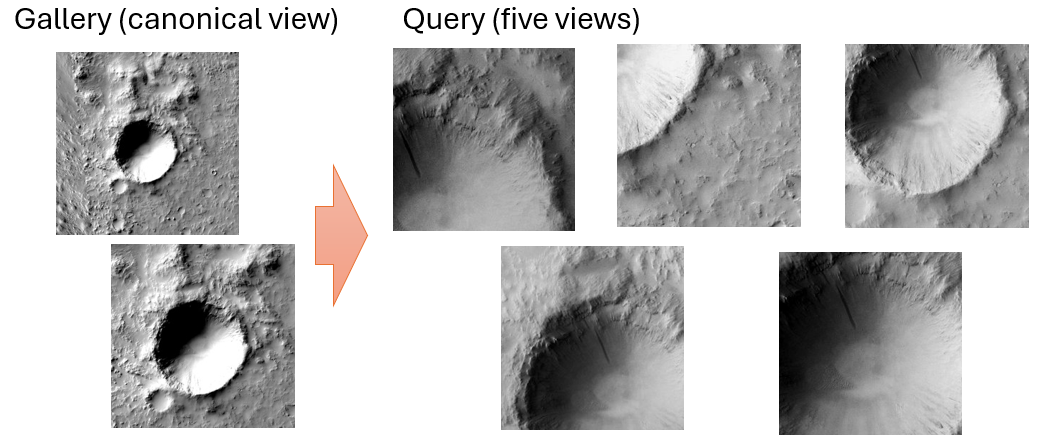}
    \caption{Examples of Robbins \cite{robbins2012new} crater ID \textsc{03-1-003926} in the dataset. Two canonical view and 5 different views with adjusted lighting conditions. }
    \label{fig:dataset-view}
\end{figure}

\paragraph{Queries.}
We select 1{,}000 crater IDs as queries, each represented by five distinct views (5{,}000 query images total).
Query views are initially generated automatically but manually verified to ensure informative crater content and to exclude degenerate cases (pure background, ambiguous partial coverage, severe artifacts).
Views vary crop placement/context and apply controlled photometric adjustments (Fig.~\ref{fig:dataset-view}).
Query crater IDs are balanced across regions to avoid spatial clustering.

\paragraph{Splits.}
We partition identities into 24,000 \textit{train} craters and 1,000 \textit{query} craters.
All fine-tuning uses only train identities; query identities are held out from training and used exclusively for evaluation.
Queries are disjoint from the gallery crops via hash- and metadata-based deduplication.

\begin{table}[t]
  \centering
  \caption{Summary of the CraterBench-R benchmark.}
  \label{tab:benchmark}
  \begin{tabular}{@{}lr@{}}
    \toprule
    \textbf{Property} & \textbf{Value} \\
    \midrule
    Crater identities          & 25{,}000 \\
    Gallery images             & 50{,}000 (2$\times$ and 3$\times$ context) \\
    Query identities           & 1{,}000 \\
    Query images               & 5{,}000 (5 views each) \\
    Regions                    & 8 MC quadrangles \\
    Diameter  median    & 1.5 km \\
    Multi-ID relevance         & 6\% \\
    \bottomrule
  \end{tabular}
\end{table}

\subsection{Evaluation Metrics}
We report Recall@$K$ ($K\in\{1,5,10\}$) and mean Average Precision (mAP). For each query $q$, the relevant set $\mathcal{R}(q)\subset\mathcal{G}$ comprises all gallery images whose crater ID belongs to $\mathcal{I}(q)$. Because each gallery identity contributes exactly two images and 94\% of queries have $|\mathcal{I}(q)|{=}1$, the vast majority of queries have $|\mathcal{R}(q)|{=}2$.


%% file: sec/2b_method.tex
\section{Method}
\label{sec:aggregation}

Single-vector pooling collapses spatial structure into one descriptor, imposing a low accuracy ceiling (Sec.~\ref{sec:exp_baselines}), while retaining all $N{=}196$ patch tokens is impractical at scale.
We bridge this gap with \emph{instance-token aggregation}: a deterministic, training-free compression of $N$ tokens into $K \ll N$ instance tokens, matched via late interaction within a two-stage retrieval pipeline.

\paragraph{Late-interaction matching.}
Given L2-normalized token sets $T^q{=}\{\mathbf{t}_i^q\}_{i=1}^{K_q}$ and $T^g{=}\{\mathbf{t}_j^g\}_{j=1}^{K_g}$ for a query--gallery pair, we score similarity via ColBERT-style~\cite{khattab2020colbert} late interaction:
\begin{equation}
\label{eq:late_interaction}
s_{\mathrm{LI}}(q, g) = \frac{1}{K_q}\sum_{i=1}^{K_q}\max_{1 \le j \le K_g}\;
\langle \mathbf{t}_i^q,\, \mathbf{t}_j^g \rangle.
\end{equation}

\paragraph{Instance-token aggregation.}
Let $\mathbf{t}_i {\in} \mathbb{R}^D$ ($i {=} 1, \ldots, N$) be the L2-normalized patch embeddings from the last transformer block, and $a_i$ the CLS-to-patch attention weight for token~$i$ (mean over all heads in the final layer).
We produce $K$ instance tokens $\{\mathbf{z}_k\}_{k=1}^K$ in three steps.
\textbf{(1)~Seed selection.}
We choose $K$ seed indices $\mathcal{S} {=} \{s_1, \ldots, s_K\}$ by either \emph{attention} (top-$K$ by $a_i$, prioritizing saliency) or \emph{FPS} (farthest-point sampling in cosine space, prioritizing diversity).
\textbf{(2)~Assignment.}
Each non-seed token is assigned to its nearest seed by cosine similarity, inducing clusters $C_k {=} \{i : \argmax_{k'} \cos(\mathbf{t}_i, \mathbf{t}_{s_{k'}}) {=} k,\; i {\notin} \mathcal{S}\}$---the set of patch-token \emph{indices} assigned to seed~$s_k$.
Soft variants are evaluated in Sec.~\ref{sec:suppl_agg}; hard assignment matches or exceeds them throughout.
\textbf{(3)~Aggregation.}
Each instance token combines the seed with the mean of its assigned tokens:
\begin{equation}
\label{eq:instance_token}
\mathbf{z}_k
= \ell_2\!\left(
\mathbf{t}_{s_k}
+ \tfrac{1}{\max(|C_k|,\,\epsilon)}\textstyle\sum_{i \in C_k} \mathbf{t}_i
\right),
\end{equation}
where $\epsilon$ prevents division by zero.
The residual form retains the seed's identity even in small clusters, preserving more detail than a centroid.
Instance tokens are matched via Eq.~\ref{eq:late_interaction}.

\paragraph{Two-stage retrieval.}
For planetary-scale search, Stage~1 retrieves a top-$S$ shortlist with a single-vector FAISS~\cite{johnson2019faiss} index (CLS or GeM); Stage~2 reranks the shortlist via instance-token late interaction.
Offline aggregation is $O(NK)$ per image; online cost is $O(K^2 D)$ per reranked candidate.

%% file: sec/3_experiments.tex
\section{Experiments}
\label{sec:experiments}
\subsection{Baseline Benchmarking on CraterBench-R}
\label{sec:baseline_benchmark}

\subsubsection{Setup: Feature Extractors and Retrieval Protocols}
\label{sec:baselines}

\paragraph{Pretrained Feature Extractors.}
We begin by benchmarking the baseline capabilities of modern vision backbones on CraterBench-R
to establish the performance ceiling and identify representation bottlenecks.
We extract frozen features from 30 backbones spanning ImageNet-supervised CNNs
(ResNet, EfficientNet \cite{he2016resnet,tan2019efficientnet}),
generic self-supervised ViTs (DINO, DINOv2/v3, MAE, CLIP
\cite{caron2021dino,oquab2023dinov2,simeoni2025dinov3,he2022mae,radford2021clip}),
and domain-specific models (MarsDINO \cite{fang2026marsvit}).

\paragraph{Retrieval Protocols.}
Across all experiments we evaluate features under the following retrieval regimes:
\begin{itemize}
    \item \textbf{Single-Vector Global Search}: Standard instance retrieval where spatial features
    are collapsed into one global descriptor via CLS, Global Average Pooling (GAP),
    or Generalized Mean (GeM \cite{radenovic2017gem}).
    \item \textbf{Classical Dictionary Aggregation}: Training-free local feature aggregation baselines
    that compress spatial features using a dictionary (e.g., VLAD \cite{jegou2010vlad}, NetVLAD \cite{arandjelovic2016netvlad})
    or per-image clustering (K-means).
    \item \textbf{Dense Late Interaction}: A multi-token matching upper bound that performs
    ColBERT-style \cite{khattab2020colbert,santhanam2022colbertv2}
    maximum-similarity matching across all uncompressed ViT patch tokens.
    We analyze this regime in Sec.~\ref{sec:proposed_experiments}.
\end{itemize}

\begin{table}[t]
\centering
\caption{Frozen-backbone retrieval on CraterBench-R (best pooling per model).
Representative subset; the complete 30-model table is in the supplementary.
Best in \textbf{bold}, second-best \underline{underlined}.}
\label{tab:main}
\small
\setlength{\tabcolsep}{3.5pt}
\begin{tabular}{@{}lrlcccc@{}}
\toprule
\textbf{Model} & \textbf{Params} & \textbf{Pool} & \textbf{R@1} & \textbf{R@5} & \textbf{R@10} & \textbf{mAP} \\
\midrule
\multicolumn{7}{l}{\emph{CNN (ImageNet supervised)}} \\[1pt]
EfficientNet-B0       &   4\,M & GAP  & .150 & .214 & .250 & .248 \\
ResNet-50             &  24\,M & GeM  & .142 & .217 & .248 & .244 \\
\midrule
\multicolumn{7}{l}{\emph{Self-supervised ViT}} \\[1pt]
ViT-S/16 DINO         &  22\,M & CLS  & .273 & .360 & .402 & .420 \\
ViT-B/8 DINO          &  86\,M & GeM  & .304 & .379 & .409 & .461 \\
ViT-B/14 DINOv2       &  87\,M & Max  & .240 & .323 & .360 & .377 \\
ViT-7B/16 DINOv3$_\text{sat}$ & 6.7\,B & Max & \underline{.330} & \underline{.416} & \underline{.450} & \underline{.505} \\
\midrule
\multicolumn{7}{l}{\emph{Other pretraining}} \\[1pt]
ViT-B/16 MAE          &  86\,M & GeM  & .022 & .042 & .052 & .043 \\
ViT-B/16 CLIP         &  86\,M & GeM  & .058 & .091 & .109 & .107 \\
DeiT-B/16             &  86\,M & Max  & .187 & .267 & .303 & .303 \\
\midrule
\multicolumn{7}{l}{\emph{Domain-specific (Mars imagery)}} \\[1pt]
ViT-S/16 MarsDINO     &  22\,M & GeM  & .269 & .356 & .391 & .412 \\
ViT-B/16 MarsDINO     &  85\,M & CLS  & \textbf{.374} & \textbf{.472} & \textbf{.503} & \textbf{.553} \\
\bottomrule
\end{tabular}
\end{table}

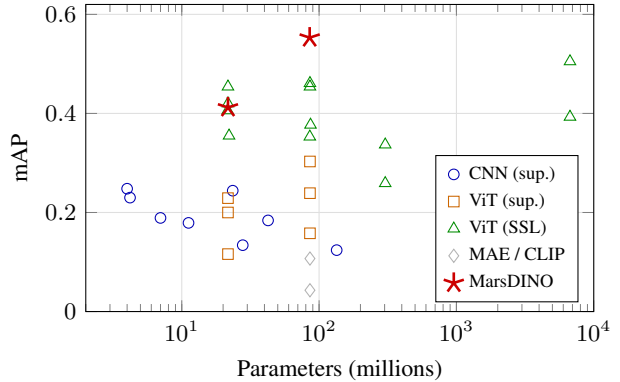
\begin{figure}[t]
\centering
\begin{tikzpicture}
\begin{semilogxaxis}[
    width=\columnwidth,
    height=0.68\columnwidth,
    xlabel={Parameters (millions)},
    ylabel={mAP},
    xmin=2, xmax=10000,
    ymin=0, ymax=0.62,
    legend pos=south east,
    legend style={font=\scriptsize, cells={anchor=west}, row sep=0pt},
    grid=major,
    grid style={gray!25},
    tick label style={font=\small},
    label style={font=\small},
]

\addplot[only marks, mark=o, mark size=2pt, blue!70!black, fill=blue!25]
  coordinates {
    (11.2,0.179) (23.5,0.244) (42.5,0.184) (27.8,0.134)
    (4.0,0.248) (7.0,0.189) (134.3,0.124) (4.2,0.230)
  };
\addlegendentry{CNN (sup.)}

\addplot[only marks, mark=square, mark size=2pt, orange!80!black, fill=orange!25]
  coordinates {
    (21.7,0.200) (85.8,0.158) (21.7,0.229) (85.8,0.303)
    (21.7,0.116) (85.8,0.239)
  };
\addlegendentry{ViT (sup.)}

\addplot[only marks, mark=triangle, mark size=2.5pt, green!55!black, fill=green!15]
  coordinates {
    (21.7,0.420) (21.7,0.454) (85.8,0.454) (85.8,0.461)
    (22.1,0.355) (86.6,0.377) (21.6,0.406) (85.6,0.353)
    (303.1,0.337) (303.1,0.259) (6716,0.393) (6716,0.505)
  };
\addlegendentry{ViT (SSL)}

\addplot[only marks, mark=diamond, mark size=2.5pt, gray!60, fill=gray!15]
  coordinates {(85.8,0.043) (85.8,0.107)};
\addlegendentry{MAE / CLIP}

\addplot[only marks, mark=star, mark size=4pt,
         red!80!black, mark options={solid, line width=1pt}]
  coordinates {(21.7,0.412) (85.4,0.553)};
\addlegendentry{MarsDINO}

\end{semilogxaxis}
\end{tikzpicture}
\caption{Model size vs.\ mAP across pretraining paradigms (all 30 backbones).
MarsDINO (\textcolor{red!80!black}{$\star$}) achieves the highest mAP with 85\,M parameters, outperforming models with up to 79$\times$ more parameters.}
\label{fig:params_vs_map}
\end{figure}

\subsubsection{Frozen Single-Vector Retrieval: The Global-Descriptor Ceiling}
\label{sec:exp_baselines}

\paragraph{Self-supervised ViTs dominate.} Table~\ref{tab:main} summarizes representative frozen-backbone results;
Fig.~\ref{fig:params_vs_map} plots all backbones evaluated.
DINO-family models outperform all CNN and supervised ViT baselines by a wide margin.
The best generic SSL model, ViT-B/8 DINO (R@1\,=\,.304), exceeds the best CNN
(EfficientNet-B0, .150) by $2\times$ and the best supervised ViT (DeiT-B/16, .187) by 63\%.
Even 22\,M-parameter ViT-S/16 DINO (.273) surpasses every supervised baseline including 86\,M variants,
indicating that self-supervised objectives~\cite{caron2021dino} learn representations better suited
to instance matching under the  domain shift.

\paragraph{Domain pretraining as a major factor.}
ViT-B/16 MarsDINO achieves the highest performance (R@1\,=\,.374, mAP\,=\,.553),
outperforming the architecturally identical ViT-B/16 DINO by +7.9 R@1 and +9.9 mAP---a gain consistent
with the benefit of in-domain pretraining.
The largest generic model, ViT-7B/16 DINOv3$_{\text{sat}}$, reaches R@1\,=\,.330 but still falls short
of the 85\,M-parameter MarsDINO, underscoring diminishing returns from generic scaling on out-of-domain retrieval.
Within DINOv3, satellite-pretrained variants (``sat'') consistently outperform the larger generic corpus (``lvd'')
at the same scale, further supporting domain proximity over data volume.

\paragraph{Pretraining objective matters more than architecture.}
ViT-B/16 MAE (.022) and CLIP (.058) both perform poorly despite sharing the same architecture as strong DINO baselines.
MAE's reconstruction objective forces the model to focus on high-frequency pixel-level reconstruction rather than learning spatially semantic representations, causing it to fail dramatically at instance-level discriminative matching (a ${>}20\%$ gap in R@1 compared to contrastive/self-distillation objectives like DINO that inherently optimize for global instance consistency). CLIP's language-aligned features
do not transfer to planetary surfaces.
Among CNNs, EfficientNet-B0 (4\,M) is the strongest (.150), outperforming much larger ResNets and VGG-16
(134\,M, .068)---parameter count alone is a poor predictor of retrieval quality across all paradigms
(Fig.~\ref{fig:params_vs_map}).

\subsubsection{Fine-Tuning vs.\ Frozen Features}
\label{sec:finetuning}

To test whether supervised metric learning can outperform frozen representations, we fine-tune ViT-S/16 MarsDINO
with three standard losses: supervised contrastive (SupCon)~\cite{khosla2020supcon}, ArcFace~\cite{deng2019arcface},
and batch-hard triplet~\cite{hermans2017defense}.
We fine-tune the full backbone with AdamW for 30 epochs (backbone lr$\,{=}\,$$10^{-5}$)
on the gallery training split (24K craters, 48K images), excluding crater identities used in the evaluation ground truth.
We evaluate pooled single-vector retrieval (CLS, GeM) and token-level late interaction (LI, $K{=}32$).
We compare two augmentation regimes: single-crop (1$\times$) and a simple multi-crop (MC) variant.

Across all configurations, fine-tuning underperforms frozen features.
With single-crop, triplet is strongest yet still reduces CLS mAP from .368 to .318 and LI from .602 to .530;
SupCon and ArcFace degrade more substantially.
Multi-crop further worsens SupCon and ArcFace and yields only a modest recovery for triplet, which remains below frozen.
Importantly, LI drops in every case, suggesting that under this low-view regime full-backbone fine-tuning disrupts the
token-level structure that late interaction exploits.
We attribute this primarily to positive-view scarcity---each crater has only two source images---rather than an inherent conflict between metric learning and patch-level representations.
Richer multi-view training data or token-aware objectives may eventually succeed.

\paragraph{Takeaway.}
Baseline benchmarking indicates that (i) global pooling imposes a strong ceiling even with strong backbones,
and (ii) with the limited views available, supervised fine-tuning does not improve---and often degrades---retrieval.
We therefore keep backbones frozen and focus on \emph{training-free token-level matching and compression}.

\subsection{Proposed Method Experiments}
\label{sec:proposed_experiments}

\subsubsection{Dense Multi-Token Matching}
\label{sec:token}

We retain the top-$K$ patch tokens (out of 196) from the last transformer block and retrieve via late interaction
(Eq.~\ref{eq:late_interaction}).
We compare two token selection strategies: \textbf{attention} (top-$K$ by CLS$\to$patch attention)
and \textbf{random} (uniform sample), evaluated on ViT-S/16 with both DINO and MarsDINO weights.

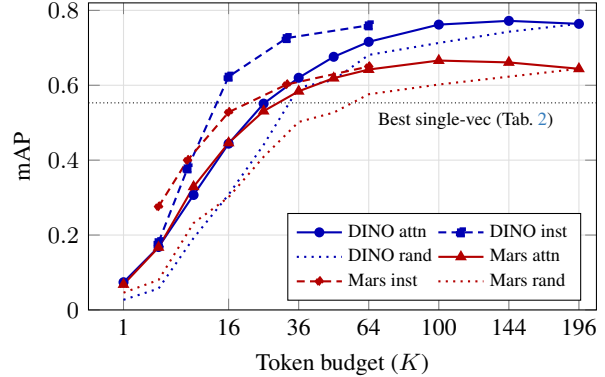
\begin{figure}[t]
\centering
\begin{tikzpicture}
\begin{axis}[
    width=\columnwidth,
    height=0.68\columnwidth,
    xlabel={Token budget ($K$)},
    ylabel={mAP},
    xmin=0, xmax=14.5,
    ymin=0, ymax=0.82,
    xtick={1,4,6,8,10,12,14},
    xticklabels={1,16,36,64,100,144,196},
    legend pos=south east,
    legend style={font=\scriptsize, cells={anchor=west},
                  row sep=-1pt, legend columns=2},
    grid=major,
    grid style={gray!25},
    tick label style={font=\small},
    label style={font=\small},
]

\addplot[black, densely dotted, thin, forget plot]
  coordinates {(0,0.553) (14.5,0.553)};
\node[font=\scriptsize, anchor=south west] at (axis cs:8,0.458)
  {Best single-vec (Tab.~\ref{tab:main})};

\addplot[blue!70!black, thick, mark=*, mark size=1.5pt]
  coordinates {
    (1,0.074)(2,0.168)(3,0.307)(4,0.444)(5,0.551)
    (6,0.620)(7,0.676)(8,0.716)(10,0.762)(12,0.772)(14,0.764)
  };
\addlegendentry{DINO attn}

\addplot[blue!70!black, thick, densely dashed, mark=square*, mark size=1.5pt]
  coordinates {
    (2,0.181)(2.828,0.378)(4,0.623)(5.657,0.726)(8,0.760)
  };
\addlegendentry{DINO inst}

\addplot[blue!70!black, thick, dotted, mark=none]
  coordinates {
    (1,0.027)(2,0.057)(3,0.192)(4,0.308)(5,0.441)
    (6,0.594)(7,0.614)(8,0.681)(10,0.713)(12,0.743)(14,0.764)
  };
\addlegendentry{DINO rand}

\addplot[red!70!black, thick, mark=triangle*, mark size=1.8pt]
  coordinates {
    (1,0.068)(2,0.167)(3,0.329)(4,0.446)(5,0.531)
    (6,0.584)(7,0.619)(8,0.642)(10,0.666)(12,0.661)(14,0.644)
  };
\addlegendentry{Mars attn}

\addplot[red!70!black, thick, densely dashed, mark=diamond*, mark size=1.8pt]
  coordinates {
    (2,0.276)(2.828,0.400)(4,0.528)(5.657,0.602)(8,0.650)
  };
\addlegendentry{Mars inst}

\addplot[red!70!black, thick, dotted, mark=none]
  coordinates {
    (1,0.046)(2,0.080)(3,0.232)(4,0.301)(5,0.409)
    (6,0.502)(7,0.527)(8,0.576)(10,0.602)(12,0.623)(14,0.644)
  };
\addlegendentry{Mars rand}

\end{axis}
\end{tikzpicture}
\caption{Retrieval quality vs.\ token budget ($K$) on ViT-S/16.
Solid: raw attention-selected tokens; dashed: instance tokens
(Sec.~\ref{sec:aggregation}); dotted: random.
At $K{=}16$, instance-token aggregation lifts DINO mAP from .444
to .623 (+18\,pts).
Dotted horizontal line: best single-vector baseline
(Tab.~\ref{tab:main}).}
\label{fig:token_ablation}
\end{figure}

\paragraph{Multi-token retrieval dramatically outperforms single-vector pooling.}
Figure~\ref{fig:token_ablation} shows mAP as a function of retained tokens.
With just 64 attention-selected tokens, ViT-S/16 DINO reaches mAP\,=\,.716---surpassing even the best
single-vector baseline from Table~\ref{tab:main} (ViT-B/16 MarsDINO, mAP\,=\,.553) by +16.3 points,
despite having 4$\times$ fewer parameters and no domain-specific pretraining.
MarsDINO shows the same pattern (mAP\,=\,.642 at $K{=}64$).
Performance plateaus around $K{\in}[64, 100]$. Retaining all 196 tokens slightly \emph{hurts} MarsDINO
(mAP drops from .666 to .644), suggesting that uninformative background tokens add noise to the matching.

\paragraph{Attention-based selection outperforms random.}
Selecting tokens by CLS$\to$patch attention consistently beats random sampling, with the largest gap at low $K$:
at $K{=}16$, attention selection leads random by +14 mAP on DINO and +15 on MarsDINO.
The gap narrows as $K$ grows, confirming that the benefit of selection is in \emph{prioritizing} informative tokens
when the budget is tight.

\subsubsection{Ablations: Instance-Token Aggregation}
\label{sec:exp_aggregation}

\paragraph{Aggregation improves the token-budget/accuracy tradeoff.}
Figure~\ref{fig:token_ablation} (dashed curves) shows that instance tokens consistently outperform raw selected tokens
at every $K$.
The gains are largest at small budgets: at $K{=}16$, DINO instance tokens reach mAP\,=\,.623---a +17.9-point jump over
raw attention selection (.444)---comparable to increasing raw attention-selected tokens from $K{=}16$ to roughly
$K{\approx}36$.
At $K{=}64$, aggregation pushes DINO from .716 to .760.
MarsDINO shows consistent gains: $K{=}16$ improves from .446 to .528, and $K{=}64$ reaches .650, slightly exceeding
the full 196-token baseline (.644) with $3{\times}$ less storage.

\paragraph{Domain-specific attention informs seed selection.}
The divergence between DINO and MarsDINO on seed strategy is highly instructive.
DINO, trained on ImageNet, develops attention heads that focus on generic foreground objects; FPS seeds compensate
by enforcing spatial diversity across the token space (mAP .760 vs.\ .734 at $K{=}64$).
Conversely, MarsDINO, pretrained on Mars imagery, has attention heads inherently calibrated to crater-relevant regions,
making attention-based seeds naturally sufficient (.650 vs.\ .632).
Additionally, hard nearest-neighbor assignment performs on par with soft variants while being simpler to compute.
Late interaction remains essential: pooling instance tokens into one vector recovers only 40--60\% of the LI mAP.

\subsubsection{Comparison vs.\ Classical Clustering}
\label{sec:vlad_comparison}

To compare instance tokens to classical local feature aggregation, we evaluate VLAD~\cite{jegou2010vlad} and NetVLAD~\cite{arandjelovic2016netvlad} on the same frozen ViT-S/16 patch tokens.
Both methods cluster the 196 patch tokens into $K$ groups via K-means and accumulate per-cluster residuals; NetVLAD uses soft assignment.
We evaluate in two modes at matched byte budgets: (i)~single-vector search after PCA whitening to 384\,d (1,536\,B, matching CLS), and (ii)~multi-vector late interaction with $K$ cluster descriptors ($K{\times}384{\times}4$\,B).

\begin{table}[t]
\centering
\caption{Comparison with aggregation baselines on ViT-S/16 at matched byte budgets.
SV\,=\,single-vector (cosine); LI\,=\,late interaction.
Per-image K-means leads at $K{=}16$; instance tokens dominate at $K{=}64$.}
\label{tab:vlad_comparison}
\small
\setlength{\tabcolsep}{3pt}
\begin{tabular}{@{}ll r cc cc@{}}
\toprule
& & & \multicolumn{2}{c}{\textbf{DINO}} & \multicolumn{2}{c}{\textbf{MarsDINO}} \\
\cmidrule(lr){4-5} \cmidrule(lr){6-7}
\textbf{Method} & \textbf{Ret.} & \textbf{B/img} & \textbf{R@1} & \textbf{mAP} & \textbf{R@1} & \textbf{mAP} \\
\midrule
\multicolumn{7}{l}{\emph{Single vector (1,536\,B)}} \\[1pt]
Best pool (Tab.~\ref{tab:main}) & SV & 1,536 & .273 & .420 & .269 & .412 \\
NetVLAD+PCA          & SV & 1,536 & .286 & .437 & .269 & .413 \\
\midrule
\multicolumn{7}{l}{\emph{Late interaction, $K{=}16$ (24\,KB)}} \\[1pt]
VLAD (global)        & LI & 24,576 & .307 & .459 & .298 & .452 \\
Per-image K-means    & LI & 24,576 & \textbf{.448} & \textbf{.651} & \textbf{.362} & \textbf{.537} \\
Inst.\ tokens (ours) & LI & 24,576 & .425 & .623 & .353 & .528 \\
\midrule
\multicolumn{7}{l}{\emph{Late interaction, $K{=}64$ (98\,KB)}} \\[1pt]
VLAD (global)        & LI & 98,304 & .377 & .551 & .334 & .497 \\
Per-image K-means    & LI & 98,304 & .530 & .746 & .429 & .615 \\
Inst.\ tokens (ours) & LI & 98,304 & \textbf{.541} & \textbf{.760} & \textbf{.457} & \textbf{.650} \\
\bottomrule
\end{tabular}
\end{table}

\begin{figure}[t]
\centering
\begin{tikzpicture}
\begin{semilogxaxis}[
    width=\columnwidth,
    height=0.72\columnwidth,
    xlabel={Bytes per image},
    ylabel={mAP},
    xmin=1000, xmax=140000,
    ymin=0.36, ymax=0.80,
    xtick={1536, 24576, 49152, 98304},
    xticklabels={1.5K, 24K, 49K, 98K},
    legend pos=north west,
    legend style={font=\scriptsize, cells={anchor=west},
                  row sep=-2pt, legend columns=1},
    grid=major,
    grid style={gray!25},
    tick label style={font=\small},
    label style={font=\small},
]

\addplot[blue!70!black, thick, mark=*, mark size=1.8pt]
  coordinates {(24576,0.623)(49152,0.726)(98304,0.760)};
\addlegendentry{DINO inst (ours)}

\addplot[blue!70!black, thick, dashdotted, mark=x, mark size=2.5pt]
  coordinates {(24576,0.651)(49152,0.709)(98304,0.746)};
\addlegendentry{DINO K-means}

\addplot[blue!70!black, thick, densely dashed, mark=square*, mark size=1.5pt]
  coordinates {(24576,0.459)(49152,0.520)(98304,0.551)};
\addlegendentry{DINO VLAD}

\addplot[red!70!black, thick, mark=triangle*, mark size=2pt]
  coordinates {(24576,0.528)(49152,0.602)(98304,0.650)};
\addlegendentry{Mars inst (ours)}

\addplot[red!70!black, thick, dashdotted, mark=+, mark size=2.5pt]
  coordinates {(24576,0.537)(49152,0.583)(98304,0.615)};
\addlegendentry{Mars K-means}

\addplot[red!70!black, thick, densely dashed, mark=diamond*, mark size=2pt]
  coordinates {(24576,0.452)(49152,0.474)(98304,0.497)};
\addlegendentry{Mars VLAD}

\addplot[only marks, mark=*, mark size=2pt, blue!70!black, forget plot]
  coordinates {(1536,0.420)(1536,0.437)};
\node[font=\tiny, anchor=south west, blue!70!black]
  at (axis cs:1700,0.438) {NV};
\node[font=\tiny, anchor=north west, blue!70!black]
  at (axis cs:1700,0.418) {CLS};
\addplot[only marks, mark=triangle*, mark size=2pt, red!70!black, forget plot]
  coordinates {(1536,0.4125)};
\node[font=\tiny, anchor=north west, red!70!black]
  at (axis cs:1700,0.406) {GeM/NV};

\node[font=\tiny, blue!70!black, anchor=south west]
  at (axis cs:25500,0.625) {\textit{16}};
\node[font=\tiny, blue!70!black, anchor=south west]
  at (axis cs:51000,0.728) {\textit{32}};
\node[font=\tiny, blue!70!black, anchor=south west]
  at (axis cs:102000,0.762) {\textit{64}};

\end{semilogxaxis}
\end{tikzpicture}
\caption{mAP vs.\ storage budget (bytes/image) on ViT-S/16.
Solid: instance tokens (ours); dash-dotted: per-image K-means;
dashed: VLAD (global centroids); points at 1.5\,KB: single-vector baselines.
Per-image K-means starts ahead at $K{=}16$ but instance tokens dominate at $K{\geq}32$.}
\label{fig:map_vs_bytes}
\end{figure}
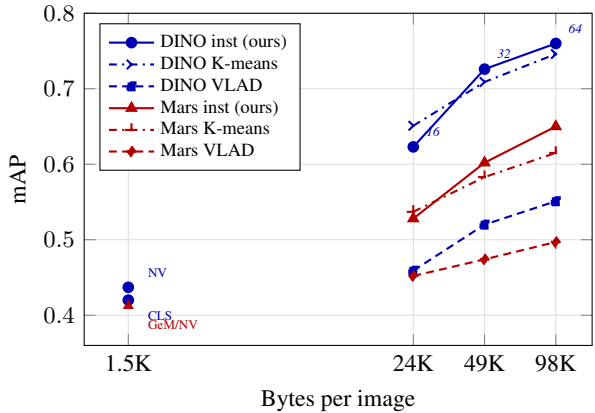

Table~\ref{tab:vlad_comparison} and Fig.~\ref{fig:map_vs_bytes} summarize the results.
As a single vector, NetVLAD+PCA slightly exceeds CLS on DINO (.437 vs.\ .420 mAP) and matches GeM on MarsDINO (.413 vs.\ .412), confirming that classical aggregation is a competitive pooling strategy.

\paragraph{Per-image clustering is critical.}
In the multi-vector regime, both per-image methods---K-means and instance tokens---dramatically outperform global VLAD at every budget.
At $K{=}16$, DINO per-image K-means reaches mAP\,=\,.651 vs.\ VLAD's .459 (+19.2\,pts); at $K{=}64$, instance tokens reach .760 vs.\ .551 (+20.9\,pts).
This  shows that per-image token grouping successfully captures localized discriminative structure that global centroids wash out.

\paragraph{Instance tokens vs.\ per-image K-means.}
Per-image K-means is the most conceptually similar baseline to our method.
At $K{=}16$, K-means centroids (iterated cluster means) slightly outperform instance tokens on both backbones (DINO mAP .651 vs.\ .623; MarsDINO .537 vs.\ .528), because with ${\sim}$12 tokens per cluster the centroid is a near-optimal representative.
However, instance tokens overtake K-means at the operationally relevant budgets of $K{\geq}32$ (DINO .726 vs.\ .709) and extend the lead at $K{=}64$ (DINO .760 vs.\ .746; MarsDINO .650 vs.\ .615).
As $K$ grows, clusters shrink to ${\sim}$3 tokens and our seed-plus-mean formula---which preserves the identity of the physical seed token rather than collapsing to a smoothed spatial average---retains more discriminative detail than a centroid.
Replacing centroids with medoids (actual tokens closest to the cluster mean) degrades further (DINO mAP .488 at $K{=}16$, .712 at $K{=}64$), confirming our token averaging is essential.
Beyond accuracy, instance tokens are vastly simpler for system design: they require no iterative optimization (single-pass assignment vs.\ 20 Lloyd iterations) and produce fully deterministic descriptors.


\subsubsection{Planetary-Scale Deployment: Two-Stage Retrieval}
\label{sec:two_stage}

Late interaction over the full gallery scales as $\mathcal{O}(K^2 D \cdot N_g)$ per query, which is prohibitive at a planetary scale.
A highly practical alternative is two-stage retrieval: (1)~single-vector FAISS search retrieves a top-$S$ shortlist, and (2)~late interaction with instance tokens reranks only the shortlist.

\begin{table}[t]
\centering
\caption{Two-stage retrieval with $K{=}32$ instance tokens at varying shortlist sizes $S$.
R@$S$: shortlist recall (accuracy ceiling for reranking).
Full LI: exhaustive late interaction over the entire 50K gallery.
Time = mean per-query latency over the 5K queries (FAISS search + reranking; offline aggregation excluded).}
\label{tab:two_stage}
\small
\setlength{\tabcolsep}{4pt}
\begin{tabular}{@{}l r ccc r@{}}
\toprule
& {$S$} & {R@1} & {mAP} & {R@$S$} & {Time (ms)} \\
\midrule
\multicolumn{6}{l}{\emph{ViT-S/16 DINO (CLS shortlist, FPS/hard seeds)}} \\[1pt]
\quad Stage 1 only  &  ---  & .273 & .422 &  ---  & 0.6 \\
\quad +Rerank       &   50  & .446 & .614 & .497 & 2.0 \\
\quad +Rerank       &  100  & .463 & .644 & .534 & 3.6 \\
\quad +Rerank       &  200  & .478 & .668 & .576 & 6.4 \\
\quad +Rerank       &  500  & .497 & .695 & .629 & 15.0 \\
\quad Full LI       &   --- & .514 & .726 & ---  &  --- \\
\midrule
\multicolumn{6}{l}{\emph{ViT-S/16 MarsDINO (GeM shortlist, Attn/hard seeds)}} \\[1pt]
\quad Stage 1 only  &  ---  & .269 & .413 &  ---  & 0.6 \\
\quad +Rerank       &   50  & .389 & .548 & .472 & 2.0 \\
\quad +Rerank       &  100  & .397 & .566 & .507 & 3.4 \\
\quad +Rerank       &  200  & .404 & .580 & .542 & 6.4 \\
\quad +Rerank       &  500  & .410 & .591 & .586 & 15.0 \\
\quad Full LI       &   --- & .415 & .602 & ---  &  --- \\
\bottomrule
\end{tabular}
\end{table}

\begin{figure*}[ht]
\centering
\includegraphics[width=\textwidth]{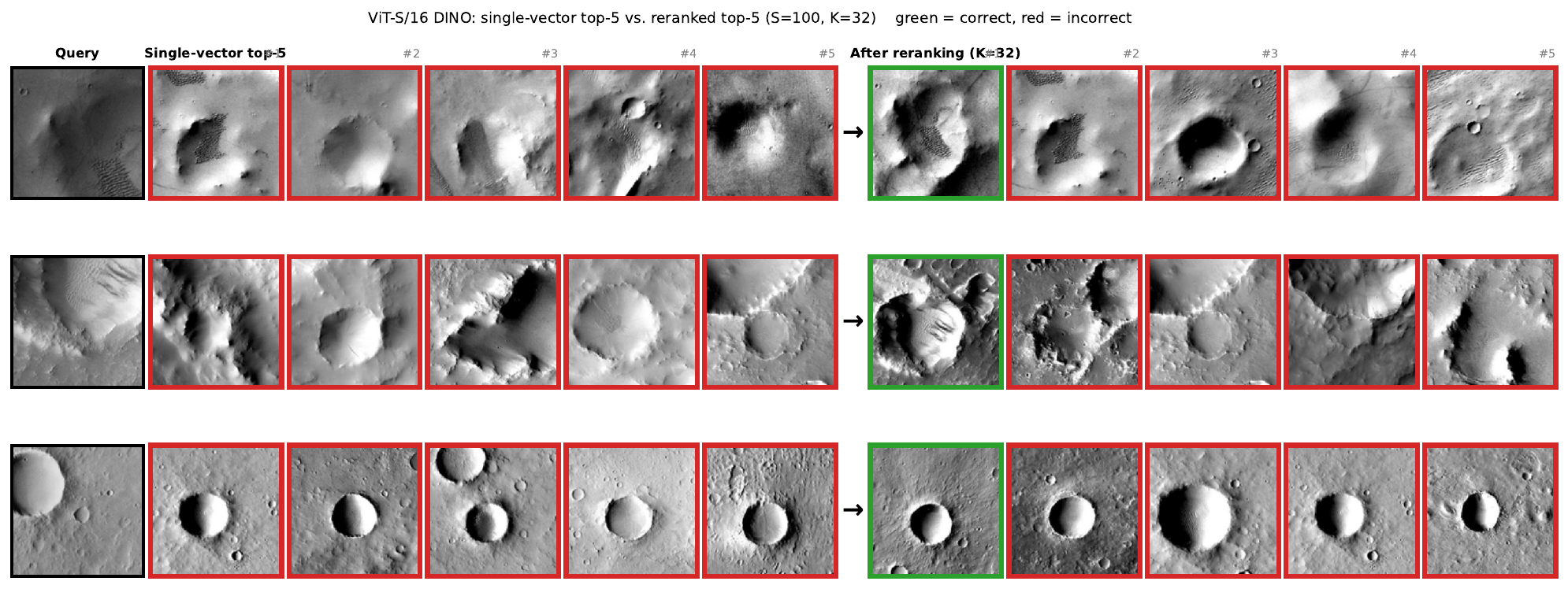}
\caption{Qualitative two-stage retrieval on ViT-S/16 DINO ($S{=}100$, $K{=}32$).
Each row shows a query and its top-5 retrieved images before (single-vector) and after (instance-token reranking).
\textcolor{green!50!black}{Green} = correct match, \textcolor{red}{red} = incorrect.
In all three examples, the correct match is absent from the single-vector top-5 (ranked 98--99) but jumps to rank 1 after reranking.}
\label{fig:qualitative_two_stage}
\end{figure*}

Table~\ref{tab:two_stage} sweeps shortlist sizes $S\in\{50,100,200,500\}$ with $K{=}32$ instance tokens.
Stage~1 FAISS search over the 50K gallery takes ${\sim}$3\,s for the full 5K-query evaluation set (about 0.6\,ms/query), largely independent of $S$.
Shortlist recall (R@$S$) grows from 50\% at $S{=}50$ to 63\% at $S{=}500$ for DINO, setting the accuracy ceiling for reranking.
The two-stage results confirm the practical viability of this pipeline: shortlist size $S$ provides a direct accuracy/latency knob. Even at an aggressive $S{=}50$ (2.0\,ms/query; $\sim$10\,s for 5K queries), reranking lifts DINO mAP from .422 to .614 (+19.2 points), recovering 85--91\% of exhaustive mAP.
Increasing $S$ to 500 (15.0\,ms/query; $\sim$75\,s for 5K queries) pushes mAP to .695 (96\% of the exhaustive LI bound).

Finally, instance tokens are highly robust to downstream quantization: INT8 quantization ($4{\times}$ compression) shows no measurable drop in mAP, and product quantization at $16{\times}$ loses only ${\sim}$1 mAP point (Table~\ref{tab:suppl-quant} in supplementary). Coupled with the two-stage FAISS pipeline, KB-scale per-image storage with rapid query latencies establishes a highly viable operational threshold for practitioners.
Figure~\ref{fig:qualitative_two_stage} visually illustrates this deployment effect: queries whose correct match is ranked near the bottom of the shortlist by single-vector search are aggressively promoted to rank~1 after instance-token reranking.

\section{Conclusion}

Our extensive evaluation reveals a fundamental hierarchy of factors for scaling planetary crater retrieval: \emph{matching strategy (late interaction $\gg$ pooling) $>$ token aggregation $>$ pretraining objective $>$ architecture scale}. The largest accuracy gains stem not from expanding parameter counts or supervised fine-tuning, but from properly engineering how spatial token representations are matched at query time.

By formulating crater analysis as instance-level retrieval, we introduced CraterBench-R and demonstrated that standard single-vector pooling inherently destroys structural geological identity. Because view-starvation in planetary data prevents standard supervised metric learning from fixing this representation gap, our post-hoc Instance-Token Aggregation uniquely bridges the divide. Our pipeline achieves near-dense accuracy at $3{\times}$ to $6{\times}$ less storage, cleanly outperforming classical  clustering at operationally relevant token budgets. Packaged within a two-stage retrieval pipeline, practitioners can recover 85--98\% of full accuracy with millisecond-scale per-query latency, making ColBERT-style dense retrieval viable for planetary-scale catalogs.

Several directions remain open. Richer training data with more varying illumination views per identity may eventually enable effective metric learning where current galleries fall short. Additionally, integrating ColBERT-style ANN indexing~\cite{santhanam2022colbertv2} for direct multi-vector search could theoretically replace the two-stage paradigm entirely, unlocking sub-linear queries. Finally, extending the benchmark to the Moon, Mercury, and other bodies will test the true cross-domain generality of these foundational representations.

%% file: sec/X_suppl.tex
\clearpage
\setcounter{page}{1}
\maketitlesupplementary

\section{Complete Baseline Results}
\label{sec:suppl_full}

Table~\ref{tab:full_baselines} reports retrieval performance for all 30 frozen backbones evaluated on Curated-5K, using the best pooling strategy per model.
This extends Table~\ref{tab:main} in the main paper, which shows a representative subset.

\begin{table*}[t]
\centering
\caption{Complete frozen-backbone retrieval results on Curated-5K (best pooling per model). Best per column in \textbf{bold}, second-best \underline{underlined}.}
\label{tab:full_baselines}
\small
\setlength{\tabcolsep}{5pt}
\begin{tabular}{@{}llrlcccc@{}}
\toprule
\textbf{Model} & \textbf{Pretraining} & \textbf{Params} & \textbf{Pool} & \textbf{R@1} & \textbf{R@5} & \textbf{R@10} & \textbf{mAP} \\
\midrule
\multicolumn{8}{l}{\emph{CNNs (ImageNet-1k supervised)}} \\[2pt]
ResNet-18          & Sup.\ IN-1k       &  11\,M & GAP  & .102 & .155 & .182 & .179 \\
ResNet-50          & Sup.\ IN-1k       &  24\,M & GeM  & .142 & .217 & .248 & .244 \\
ResNet-101         & Sup.\ IN-1k       &  43\,M & GAP  & .103 & .170 & .200 & .184 \\
ConvNeXt-T         & Sup.\ IN-1k       &  28\,M & GeM  & .075 & .121 & .146 & .134 \\
EfficientNet-B0    & Sup.\ IN-1k       &   4\,M & GAP  & .150 & .214 & .250 & .248 \\
DenseNet-121       & Sup.\ IN-1k       &   7\,M & GAP  & .110 & .170 & .202 & .189 \\
VGG-16             & Sup.\ IN-1k       & 134\,M & GAP  & .068 & .108 & .134 & .124 \\
MobileNetV3-L      & Sup.\ IN-1k       &   4\,M & GAP  & .133 & .201 & .235 & .230 \\
\midrule
\multicolumn{8}{l}{\emph{Supervised ViTs}} \\[2pt]
ViT-S/16 AugReg    & Sup.\ IN-21k${\to}$1k  &  22\,M & Mean & .115 & .187 & .219 & .200 \\
ViT-B/16 Orig      & Sup.\ IN-21k${\to}$1k  &  86\,M & Mean & .091 & .142 & .171 & .158 \\
DeiT-S/16          & Sup.\ IN-1k       &  22\,M & GeM  & .137 & .202 & .232 & .229 \\
DeiT-B/16          & Sup.\ IN-1k       &  86\,M & Max  & .187 & .267 & .303 & .303 \\
DeiT3-S/16         & Sup.\ IN-1k       &  22\,M & Mean & .064 & .109 & .129 & .116 \\
DeiT3-B/16         & Sup.\ IN-1k       &  86\,M & GeM  & .148 & .210 & .235 & .239 \\
\midrule
\multicolumn{8}{l}{\emph{Self-supervised ViTs (natural images)}} \\[2pt]
ViT-S/16 DINO      & SSL (IN-1k)       &  22\,M & CLS  & .273 & .360 & .402 & .420 \\
ViT-S/8  DINO      & SSL (IN-1k)       &  22\,M & CLS  & .296 & .383 & .419 & .454 \\
ViT-B/16 DINO      & SSL (IN-1k)       &  86\,M & CLS  & .295 & .387 & .425 & .454 \\
ViT-B/8  DINO      & SSL (IN-1k)       &  86\,M & GeM  & .304 & .379 & .409 & .461 \\
ViT-S/14 DINOv2    & SSL (LVD-142M)    &  22\,M & Max  & .226 & .302 & .336 & .355 \\
ViT-B/14 DINOv2    & SSL (LVD-142M)    &  87\,M & Max  & .240 & .323 & .360 & .377 \\
ViT-S/16 DINOv3    & SSL (LVD-1.7B)    &  22\,M & CLS  & .258 & .354 & .395 & .406 \\
ViT-B/16 DINOv3    & SSL (LVD-1.7B)    &  86\,M & CLS  & .218 & .321 & .363 & .353 \\
ViT-L/16 DINOv3$_\text{sat}$  & SSL (Sat-493M) & 303\,M & Max & .208 & .287 & .321 & .337 \\
ViT-L/16 DINOv3$_\text{lvd}$  & SSL (LVD-1.7B) & 303\,M & CLS & .162 & .233 & .274 & .259 \\
ViT-7B/16 DINOv3$_\text{lvd}$ & SSL (LVD-1.7B) & 6.7\,B & Max & .247 & .340 & .377 & .393 \\
ViT-7B/16 DINOv3$_\text{sat}$ & SSL (Sat-493M) & 6.7\,B & Max & \underline{.330} & \underline{.416} & \underline{.450} & \underline{.505} \\
\midrule
\multicolumn{8}{l}{\emph{Other pretraining objectives}} \\[2pt]
ViT-B/16 MAE       & Recon.\ (IN-1k)   &  86\,M & GeM  & .022 & .042 & .052 & .043 \\
ViT-B/16 CLIP      & Lang.-img (WIT)   &  86\,M & GeM  & .058 & .091 & .109 & .107 \\
\midrule
\multicolumn{8}{l}{\emph{Domain-specific (Mars orbital imagery)}} \\[2pt]
ViT-S/16 MarsDINO  & DINO (Mars)       &  22\,M & GeM  & .269 & .356 & .391 & .412 \\
ViT-B/16 MarsDINO  & DINO (Mars)       &  85\,M & CLS  & \textbf{.374} & \textbf{.472} & \textbf{.503} & \textbf{.553} \\
\bottomrule
\end{tabular}
\end{table*}

\section{Pooling Ablation}
\label{sec:suppl_pooling}

Table~\ref{tab:pooling_full} reports performance for every ViT backbone under all four pooling strategies.
Table~\ref{tab:cnn_full} reports CNN results with GAP and GeM where applicable.

\paragraph{Pooling preferences vary by pretraining objective.}
CLS pooling is strongest for DINO~v1 backbones, where the self-supervised objective explicitly trains the CLS token.
DINOv2 and DINOv3 favor max pooling, likely because their training distributes discriminative information more uniformly across patch tokens.
MarsDINO is mixed: ViT-S prefers GeM while ViT-B is best with CLS.

\paragraph{GeM as a robust default.}
GeM with $p{=}3$ is within 2 points of the best strategy for nearly all backbones, making it a practical default when model-specific tuning is not feasible.

\begin{table*}[t]
\centering
\caption{Effect of token pooling on ViT backbones (R@1 / mAP). Best pooling per backbone in \textbf{bold}.}
\label{tab:pooling_full}
\small
\setlength{\tabcolsep}{4.5pt}
\begin{tabular}{@{}l cc cc cc cc@{}}
\toprule
& \multicolumn{2}{c}{\textbf{CLS}} & \multicolumn{2}{c}{\textbf{Mean}} & \multicolumn{2}{c}{\textbf{Max}} & \multicolumn{2}{c}{\textbf{GeM} ($p{=}3$)} \\
\cmidrule(lr){2-3}\cmidrule(lr){4-5}\cmidrule(lr){6-7}\cmidrule(lr){8-9}
\textbf{Backbone} & R@1 & mAP & R@1 & mAP & R@1 & mAP & R@1 & mAP \\
\midrule
\multicolumn{9}{l}{\emph{DINO v1}} \\[2pt]
ViT-S/16 DINO      & \textbf{.273} & \textbf{.420} & .177 & .279 & .126 & .210 & .258 & .392 \\
ViT-S/8  DINO      & \textbf{.296} & \textbf{.454} & .162 & .257 & .110 & .185 & .185 & .286 \\
ViT-B/16 DINO      & \textbf{.295} & \textbf{.454} & .226 & .351 & .194 & .313 & .262 & .399 \\
ViT-B/8  DINO      & .303 & .465 & .263 & .402 & .226 & .356 & \textbf{.304} & \textbf{.461} \\
\midrule
\multicolumn{9}{l}{\emph{DINOv2}} \\[2pt]
ViT-S/14 DINOv2    & .200 & .327 & .177 & .287 & \textbf{.226} & \textbf{.355} & .195 & .315 \\
ViT-B/14 DINOv2    & .157 & .262 & .188 & .306 & \textbf{.240} & \textbf{.377} & .221 & .348 \\
\midrule
\multicolumn{9}{l}{\emph{DINOv3}} \\[2pt]
ViT-S/16 DINOv3                & \textbf{.258} & \textbf{.406} & .226 & .366 & .256 & .405 & .225 & .366 \\
ViT-B/16 DINOv3                & \textbf{.218} & \textbf{.353} & .188 & .307 & .207 & .329 & .194 & .313 \\
ViT-L/16 DINOv3$_\text{sat}$  & .188 & .314 & .181 & .303 & \textbf{.208} & \textbf{.337} & .204 & .334 \\
ViT-L/16 DINOv3$_\text{lvd}$  & \textbf{.162} & \textbf{.259} & .121 & .205 & .137 & .226 & .119 & .204 \\
ViT-7B/16 DINOv3$_\text{lvd}$ & .215 & .353 & .139 & .244 & \textbf{.247} & \textbf{.393} & .204 & .337 \\
ViT-7B/16 DINOv3$_\text{sat}$ & .132 & .235 & .261 & .412 & \textbf{.330} & \textbf{.505} & .298 & .460 \\
\midrule
\multicolumn{9}{l}{\emph{MarsDINO (domain-specific)}} \\[2pt]
ViT-S/16 MarsDINO  & .241 & .368 & .249 & .381 & .225 & .355 & \textbf{.269} & \textbf{.412} \\
ViT-B/16 MarsDINO  & \textbf{.374} & \textbf{.553} & .369 & .540 & .344 & .511 & .369 & .544 \\
\midrule
\multicolumn{9}{l}{\emph{Other pretraining}} \\[2pt]
ViT-B/16 MAE       & .013 & .022 & .013 & .022 & .008 & .019 & \textbf{.022} & \textbf{.043} \\
\midrule
\multicolumn{9}{l}{\emph{Supervised ViTs}} \\[2pt]
ViT-S/16 AugReg    & .104 & .181 & \textbf{.115} & \textbf{.200} & .085 & .150 & .109 & .187 \\
ViT-B/16 Orig      & .085 & .146 & \textbf{.091} & \textbf{.158} & .080 & .135 & .082 & .143 \\
ViT-B/16 CLIP      & .019 & .043 & .056 & .103 & .049 & .090 & \textbf{.058} & \textbf{.107} \\
DeiT-S/16          & .108 & .188 & .131 & .224 & .100 & .175 & \textbf{.137} & \textbf{.229} \\
DeiT-B/16          & .159 & .260 & .182 & .292 & \textbf{.187} & \textbf{.303} & .181 & .290 \\
DeiT3-S/16         & .059 & .111 & \textbf{.064} & \textbf{.116} & .011 & .026 & .051 & .092 \\
DeiT3-B/16         & .120 & .203 & .141 & .229 & .104 & .177 & \textbf{.148} & \textbf{.239} \\
\bottomrule
\end{tabular}
\end{table*}

\begin{table}[t]
\centering
\caption{CNN baseline retrieval results. Models with both GAP and GeM pooling are shown; best per model in \textbf{bold}.}
\label{tab:cnn_full}
\small
\setlength{\tabcolsep}{4pt}
\begin{tabular}{@{}llcccc@{}}
\toprule
\textbf{Model} & \textbf{Pool} & \textbf{R@1} & \textbf{R@5} & \textbf{R@10} & \textbf{mAP} \\
\midrule
ResNet-18       & GAP  & .102 & .155 & .182 & .179 \\
\midrule
ResNet-50       & GAP  & .103 & .159 & .191 & .182 \\
ResNet-50       & GeM  & \textbf{.142} & \textbf{.217} & \textbf{.248} & \textbf{.244} \\
\midrule
ResNet-101      & GAP  & .103 & .170 & .200 & .184 \\
\midrule
ConvNeXt-T      & GAP  & .072 & .120 & .143 & .130 \\
ConvNeXt-T      & GeM  & \textbf{.075} & \textbf{.121} & \textbf{.146} & \textbf{.134} \\
\midrule
EfficientNet-B0 & GAP  & \textbf{.150} & \textbf{.214} & \textbf{.250} & \textbf{.248} \\
EfficientNet-B0 & GeM  & .141 & .207 & .235 & .239 \\
\midrule
DenseNet-121    & GAP  & .110 & .170 & .202 & .189 \\
\midrule
VGG-16          & GAP  & .068 & .108 & .134 & .124 \\
\midrule
MobileNetV3-L   & GAP  & .133 & .201 & .235 & .230 \\
\bottomrule
\end{tabular}
\end{table}

\section{Token Selection Strategy Comparison}
\label{sec:suppl_strategy}

Table~\ref{tab:strategy} compares seven token selection strategies at $K{=}64$ for both ViT-S/16 backbones.
Attention-based strategies (attention, norm$\times$attention) consistently rank first.
On DINO, the top three strategies (norm$\times$attention, attention, norm) perform within 1\% mAP of each other, while on MarsDINO the attention advantage is larger ($+$9 mAP over norm).
Random selection is a surprisingly strong baseline, achieving 95\% of the optimal mAP on DINO and 90\% on MarsDINO.
CLS-distance (selecting tokens \emph{farthest} from CLS) is consistently worst, confirming that salient---not diverse---tokens drive retrieval quality.

\begin{table}[t]
\centering
\caption{Token selection strategy comparison at $K{=}64$ (ViT-S/16). Best per model in \textbf{bold}.}
\label{tab:strategy}
\small
\setlength{\tabcolsep}{3pt}
\begin{tabular}{@{}l cc cc@{}}
\toprule
& \multicolumn{2}{c}{\textbf{DINO}} & \multicolumn{2}{c}{\textbf{MarsDINO}} \\
\cmidrule(lr){2-3}\cmidrule(lr){4-5}
\textbf{Strategy} & R@1 & mAP & R@1 & mAP \\
\midrule
Attention          & .507 & .716 & \textbf{.453} & \textbf{.642} \\
Norm$\times$Attn   & \textbf{.507} & \textbf{.717} & .450 & .641 \\
Norm               & .505 & .714 & .374 & .549 \\
Random             & .477 & .681 & .395 & .576 \\
Spatial grid       & .457 & .659 & .390 & .566 \\
CLS similarity     & .439 & .626 & .352 & .512 \\
CLS distance       & .384 & .561 & .349 & .515 \\
\bottomrule
\end{tabular}
\end{table}


\section{Instance-Token Aggregation Ablation}
\label{sec:suppl_agg}

We ablate the three design axes of the instance-token aggregation pipeline introduced in Sec.~\ref{sec:aggregation}: seed selection, token-to-seed assignment, and matching strategy.
All experiments use late interaction unless otherwise noted.

\paragraph{Assignment strategy.}
Tables~\ref{tab:suppl-agg-dino} and~\ref{tab:suppl-agg-mars} compare four assignment strategies at each $K$, using the best seed method per model.
Hard nearest-neighbor (\texttt{hard\_top1}) is the strongest or within 1\% mAP of the best for nearly all $K$ on both models.
Soft assignment (\texttt{soft\_top2/4}) is competitive but adds no consistent benefit.
Dense attention-weighted assignment (\texttt{group\_dense}) underperforms at $K{\geq}16$, likely because spreading mass across all seeds dilutes cluster coherence.
The rightmost column shows the Phase~2 raw-attention baseline (no aggregation); aggregation improves over raw at every $K$ by +1.3 to +17.9 mAP.

\begin{table}[t]
\centering
\caption{Assignment comparison --- ViT-S/16 DINO (late interaction, best seed per $K$). Raw = Phase~2 attention selection without aggregation.}
\label{tab:suppl-agg-dino}
\small
\setlength{\tabcolsep}{3pt}
\begin{tabular}{@{}r cc cc cc cc c@{}}
\toprule
& \multicolumn{2}{c}{\texttt{hard\_top1}} & \multicolumn{2}{c}{\texttt{soft\_top2}}
& \multicolumn{2}{c}{\texttt{soft\_top4}} & \multicolumn{2}{c}{\texttt{group\_dense}}
& {Raw} \\
\cmidrule(lr){2-3} \cmidrule(lr){4-5} \cmidrule(lr){6-7} \cmidrule(lr){8-9}
{$K$} & {R@1} & {mAP} & {R@1} & {mAP} & {R@1} & {mAP} & {R@1} & {mAP} & {mAP} \\
\midrule
4  & .108 & .180 & .104 & .174 & .100 & .168 & \textbf{.109} & \textbf{.181} & .168 \\
8  & \textbf{.231} & \textbf{.378} & .219 & .358 & .189 & .304 & .192 & .303 & --- \\
16 & \textbf{.425} & \textbf{.623} & .416 & .612 & .394 & .582 & .296 & .457 & .444 \\
32 & \textbf{.514} & \textbf{.726} & .511 & .724 & .497 & .708 & .432 & .628 & --- \\
64 & .541 & .759 & \textbf{.543} & \textbf{.760} & .536 & .753 & .492 & .698 & .716 \\
\bottomrule
\end{tabular}
\end{table}

\begin{table}[t]
\centering
\caption{Assignment comparison --- ViT-S/16 MarsDINO (late interaction, best seed per $K$).}
\label{tab:suppl-agg-mars}
\small
\setlength{\tabcolsep}{3pt}
\begin{tabular}{@{}r cc cc cc cc c@{}}
\toprule
& \multicolumn{2}{c}{\texttt{hard\_top1}} & \multicolumn{2}{c}{\texttt{soft\_top2}}
& \multicolumn{2}{c}{\texttt{soft\_top4}} & \multicolumn{2}{c}{\texttt{gtp\_dense}}
& {Raw} \\
\cmidrule(lr){2-3} \cmidrule(lr){4-5} \cmidrule(lr){6-7} \cmidrule(lr){8-9}
{$K$} & {R@1} & {mAP} & {R@1} & {mAP} & {R@1} & {mAP} & {R@1} & {mAP} & {mAP} \\
\midrule
4  & .163 & .267 & .166 & .270 & .169 & .275 & \textbf{.171} & \textbf{.276} & .167 \\
8  & \textbf{.258} & \textbf{.400} & .257 & .399 & .255 & .394 & .251 & .388 & --- \\
16 & \textbf{.351} & \textbf{.528} & .343 & .517 & .341 & .508 & .322 & .484 & .446 \\
32 & \textbf{.415} & \textbf{.602} & .407 & .591 & .395 & .577 & .366 & .536 & --- \\
64 & \textbf{.457} & \textbf{.650} & .449 & .639 & .439 & .626 & .388 & .561 & .642 \\
\bottomrule
\end{tabular}
\end{table}

\paragraph{Seed selection.}
Table~\ref{tab:suppl-seed} compares attention and FPS seeding.
FPS dominates on DINO at $K{\geq}8$, reaching mAP\,=\,.760 vs.\ .734 for attention at $K{=}64$---FPS produces spatially diverse seeds that better partition the token space.
On MarsDINO, attention seeds are stronger at every $K$ except 16 (where FPS edges ahead by 0.4\%), suggesting that MarsDINO's attention heads already identify instance-relevant regions.

\begin{table}[t]
\centering
\caption{Seed selection comparison (late interaction, best assignment per seed). Bold = best per model and $K$.}
\label{tab:suppl-seed}
\small
\setlength{\tabcolsep}{3pt}
\begin{tabular}{@{}r cccc cccc@{}}
\toprule
& \multicolumn{4}{c}{\textbf{DINO}} & \multicolumn{4}{c}{\textbf{MarsDINO}} \\
\cmidrule(lr){2-5}\cmidrule(lr){6-9}
& \multicolumn{2}{c}{Attention} & \multicolumn{2}{c}{FPS}
& \multicolumn{2}{c}{Attention} & \multicolumn{2}{c}{FPS} \\
\cmidrule(lr){2-3}\cmidrule(lr){4-5}\cmidrule(lr){6-7}\cmidrule(lr){8-9}
{$K$} & {R@1} & {mAP} & {R@1} & {mAP}
      & {R@1} & {mAP} & {R@1} & {mAP} \\
\midrule
4  & \textbf{.109} & \textbf{.181} & .084 & .152  & \textbf{.171} & \textbf{.276} & .113 & .201 \\
8  & .203 & .326 & \textbf{.231} & \textbf{.378}  & \textbf{.258} & \textbf{.400} & .227 & .371 \\
16 & .336 & .504 & \textbf{.425} & \textbf{.623}  & .353 & .524 & \textbf{.351} & \textbf{.528} \\
32 & .449 & .647 & \textbf{.514} & \textbf{.726}  & \textbf{.415} & \textbf{.602} & .414 & .598 \\
64 & .521 & .734 & \textbf{.543} & \textbf{.760}  & \textbf{.457} & \textbf{.650} & .444 & .632 \\
\bottomrule
\end{tabular}
\end{table}

\paragraph{Matching strategy.}
Table~\ref{tab:suppl-match} shows that late interaction is essential for multi-token retrieval: pooling instance tokens into a single vector (mean or max) recovers only 40--60\% of the late-interaction mAP.

\paragraph{Descriptor compression.}
Table~\ref{tab:suppl-quant} evaluates token quantization for storage-constrained deployment.
FP16 and INT8 (per-vector symmetric scalar quantization) are \emph{lossless}: mAP changes by ${\leq}$0.02 points on both backbones at $K{=}32$.
Product quantization with $m{=}96$ sub-vectors (16$\times$ compression) loses only 1.0--1.2 mAP points; aggressive PQ-48 (32$\times$) loses 6.2--7.3 points.
These results show that INT8 storage ($K{=}32$: 12.4\,KB/image) is a practical default, and PQ-96 (3.1\,KB/image) is viable when storage is severely constrained.

\begin{table}[t]
\centering
\caption{Effect of descriptor quantization on exhaustive late-interaction mAP ($K{=}32$).
Compression is relative to FP32 (1536\,B/token).}
\label{tab:suppl-quant}
\small
\setlength{\tabcolsep}{3pt}
\begin{tabular}{@{}l r r cc@{}}
\toprule
{Method} & {B/tok} & {Comp.} & {DINO} & {MarsDINO} \\
\midrule
FP32          & 1536 &  1$\times$ & .726 & .602 \\
FP16          &  768 &  2$\times$ & .726 & .602 \\
INT8          &  388 &  4$\times$ & .726 & .602 \\
PQ ($m{=}96$) &   96 & 16$\times$ & .715 & .590 \\
PQ ($m{=}48$) &   48 & 32$\times$ & .664 & .529 \\
\bottomrule
\end{tabular}
\end{table}
However, pooled instance tokens still outperform the single-vector baselines from Table~\ref{tab:main} when $K$ is large enough ($K{\geq}32$), confirming that aggregation concentrates discriminative signal.

\begin{table}[t]
\centering
\caption{Matching strategy comparison (attention seeds, best assignment).
Descriptor size shows bytes per image (384-dim, float32).}
\label{tab:suppl-match}
\small
\setlength{\tabcolsep}{2.5pt}
\begin{tabular}{@{}r ccc ccc@{}}
\toprule
& \multicolumn{3}{c}{\textbf{DINO}} & \multicolumn{3}{c}{\textbf{MarsDINO}} \\
\cmidrule(lr){2-4}\cmidrule(lr){5-7}
{$K$} & {Late} & {Mean} & {Max} & {Late} & {Mean} & {Max} \\
\midrule
4   & .181 & .144 & .107  & .276 & \textbf{.281} & .245 \\
8   & .326 & .207 & .141  & .400 & .343 & .311 \\
16  & .504 & .267 & .193  & .524 & .376 & .362 \\
32  & .647 & .323 & .239  & .602 & .391 & .390 \\
64  & .734 & .322 & .268  & .650 & .401 & .404 \\
\bottomrule
\end{tabular}
\end{table}